\documentclass[sigconf]{acmart}
\usepackage{tabularx}
\usepackage[table]{xcolor}
\usepackage{balance}

\AtBeginDocument{%
  }

\copyrightyear{2026}
\acmYear{2026}
\setcopyright{cc}
\setcctype{by}
\acmConference[MM '26]{Proceedings of the 34th ACM International Conference on Multimedia}{November 10--14, 2026}{Rio de Janeiro, Brazil}
\acmBooktitle{Proceedings of the 34th ACM International Conference on Multimedia (MM '26), November 10--14, 2026, Rio de Janeiro, Brazil}
\acmDOI{10.1145/3767308.3835004}
\acmISBN{979-8-4007-2213-4/2026/11}
\begin{document}

\title{StateTrace: An Object-Centric Framework for Hidden-State Spatiotemporal Reasoning in Long Videos}

\author{Yu Han}
\authornote{Yu Han and Wenhao Li contributed equally to this work.
Xiu Su is the corresponding author.}
\orcid{0009-0000-2633-3960}
\affiliation{%
  \institution{University of California, San Diego}
  \city{San Diego}
  \state{California}
  \country{United States}
}
\email{yuh162@ucsd.edu}

\author{Wenhao Li}
\authornotemark[1]
\orcid{0009-0001-4010-3536}
\affiliation{%
  \institution{The University of Sydney}
  \city{Sydney}
  \country{Australia}
}
\email{li58843972@163.com}

\author{Yichao Cao}
\orcid{0000-0003-2997-4012}
\affiliation{%
  \institution{Central South University}
  \city{Changsha}
  \country{China}
}
\email{caoyichao@csu.edu.cn}

\author{Hongyan Xu}
\orcid{0000-0003-3846-5236}
\affiliation{%
  \institution{Central South University}
  \city{Changsha}
  \country{China}
}
\email{hongyanxu@csu.edu.cn}

\author{Shuo Yang}
\orcid{0000-0001-6145-0150}
\affiliation{%
  \institution{Harbin Institute of Technology (Shenzhen)}
  \city{Shenzhen}
  \country{China}
}
\email{shuoyang@hit.edu.cn}

\author{Shan You}
\orcid{0000-0003-1964-0430}
\affiliation{%
  \institution{SenseTime Research}
  \city{Beijing}
  \country{China}
}
\email{youshan@acerobotics.com}

\author{Xiu Su}
\correspondingauthor
\orcid{0000-0002-9863-5404}
\affiliation{%
  \institution{Central South University}
  \city{Changsha}
  \country{China}
}
\email{xiusu1994@csu.edu.cn}

\renewcommand{\shortauthors}{Yu Han et al.}
\begin{abstract}
Existing VLMs have achieved strong performance in video understanding, yet they struggle with long-video spatiotemporal reasoning when target objects become invisible, often mistaking ``invisible'' for ``unknown''. We define this challenge as hidden-state spatiotemporal reasoning: inferring object states during prolonged invisible intervals from context interactions. To address this, we propose \textbf{StateTrace}, a novel object-centric framework that endows VideoLLMs with an explicit mechanism for hidden state reasoning in long videos. StateTrace builds a reusable spatiotemporal state memory that organizes object trajectories, inter-object relations, and state-transition events into a structured reasoning substrate. At inference time, it retrieves question-relevant state-evolution trajectories and converts them into compact reasoning cues, enabling the model to explicitly reason about \emph{why} an object disappears, \emph{how} its state evolves while invisible, and \emph{whether} that state should persist at query time. We further build \textbf{HSR-Bench}, a diagnostic benchmark for hidden-state reasoning, containing 1,427 video-QA samples from 1,384 unique videos. Extensive experiments across multiple VideoLLMs show that StateTrace consistently improves performance on both public benchmarks and HSR-Bench(e.g., improving VideoLLaMA3 from 39.6 to 64.2 on HSR-Bench).
\end{abstract}
\begin{CCSXML}
<ccs2012>
   <concept>
       <concept_id>10010147.10010178.10010224</concept_id>
       <concept_desc>Computing methodologies~Computer vision</concept_desc>
       <concept_significance>500</concept_significance>
       </concept>
 </ccs2012>
\end{CCSXML}

\ccsdesc[500]{Computing methodologies~Computer vision}

\keywords{Video Understanding, VLM, Hidden-State Reasoning}

\maketitle

\begin{figure*}[t]
\centering
\includegraphics[width=\textwidth]{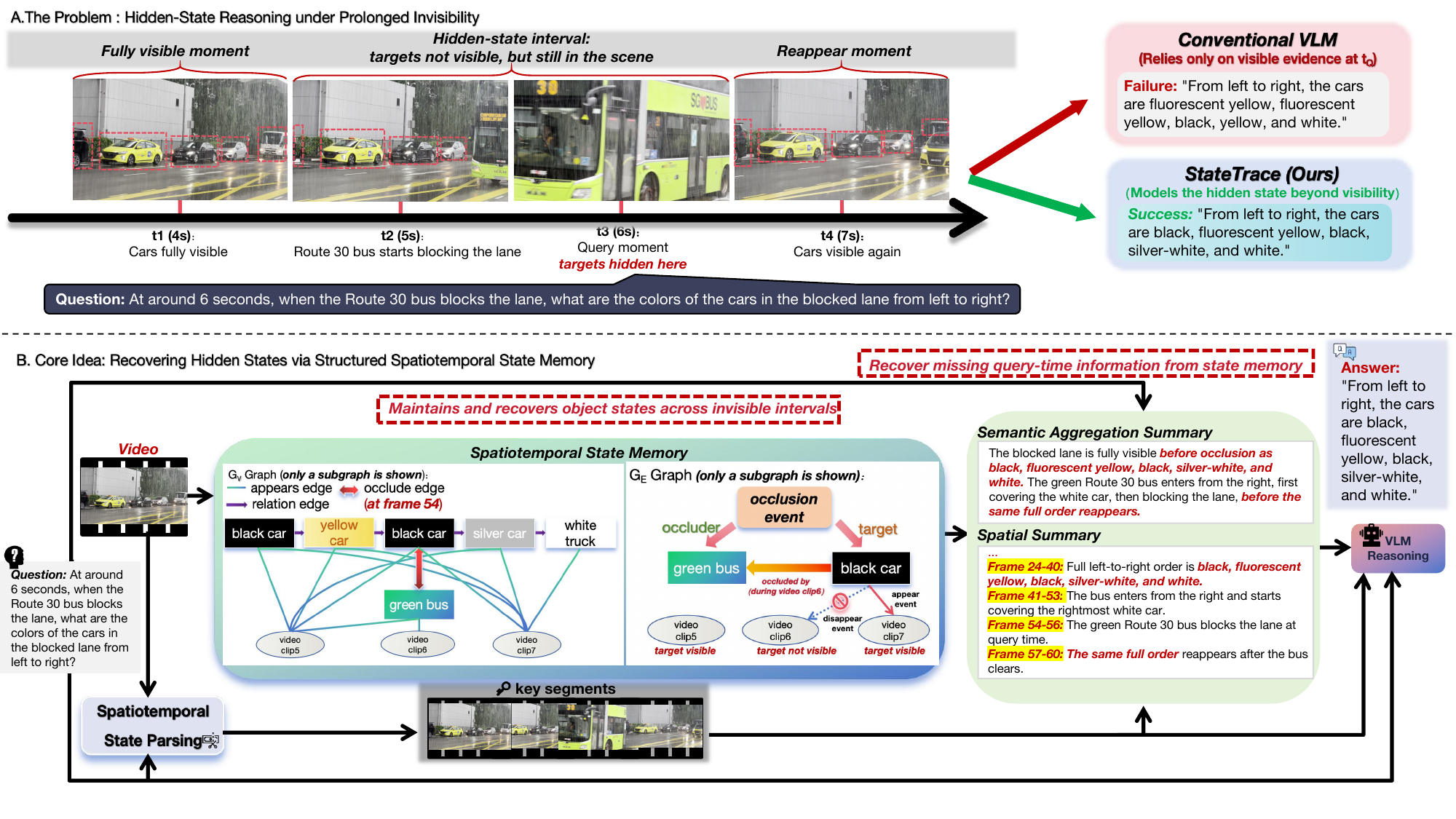}
\vspace{-33pt}
\caption{A) Illustration of hidden-state spatiotemporal reasoning; B) StateTrace: Augmented with state summaries.}
\Description{The figure contains two panels. The first panel illustrates
a target object becoming invisible in a long video. The second panel shows how StateTrace constructs and retrieves object-state summaries for answer generation.}
\vspace{-12pt}
\label{fig:core_idea}
\end{figure*}

\section{Introduction}

%In recent years, vision-language models (VLMs)~\cite{alayrac2022flamingo,li2023blip,liu2023visual,chen2024internvl} have made significant progress in open-ended video understanding~\cite{lin2024video,weng2024longvlm,maaz2024video}, and have demonstrated strong capabilities in tasks such as video captioning~\cite{yang2023vid2seq}, temporal localization~\cite{yan2023unloc}, and video question answering~\cite{maaz2024video}. However, although existing models have achieved impressive performance improvements, they still suffer from a fundamental weakness in long-video reasoning~\cite{wu2024longvideobench,wang2025lvbench,zou2025hlv}: once a target object becomes invisible in a video, the model often cannot continue to correctly infer the state of the object. For example, if a black pen is placed into a pencil case in the early part of a video and does not appear again afterward, humans can usually naturally judge that, if there is no subsequent evidence indicating that it has been taken out, then the pen is still in the pencil case at the end of the video. In contrast, existing VLMs often incorrectly answer that the pen ``disappeared'', ``is still on the table'', or ``cannot be determined''. Such errors reveal a key gap: although existing models can recognize visible content, they have difficulty reasoning about object states that persist beyond direct observation.

In recent years, vision-language models (VLMs)~\cite{alayrac2022flamingo,li2023blip,liu2023visual,chen2024internvl} have made significant progress in open-ended video understanding~\cite{lin2024video,weng2024longvlm,maaz2024video} and shown strong performance on video captioning~\cite{yang2023vid2seq}, temporal localization~\cite{yan2023unloc}, and video question answering~\cite{maaz2024video}. Yet they remain fundamentally weak at hidden-state video reasoning~\cite{wu2024longvideobench,wang2025lvbench,zou2025hlv}: once a target object becomes invisible, they often fail to infer its state. For example, if a pen is placed into a pencil case early in a video and does not appear again, then when asked where the pen is at the end of the video, humans can infer that, absent later evidence that it was taken out, it remains in the pencil case. Existing VLMs, however, often over-rely on explicit visual evidence and lack the ability to model the hidden states of temporarily invisible objects. As a result, when queried about the end of the video, they attend mainly to end-of-video visual cues and fail because the pen is not directly visible.

This limitation is especially pronounced in long videos~\cite{chandrasegaran2024hourvideo,ataallah2024infinibench}, where objects may remain invisible for extended periods due to occlusion, containment, or covering, and the query may occur long after their last visible moment. Answering such questions requires more than short-term temporal modeling~\cite{hu2024enhancing,nie2024slowfocus} or clip-level retrieval~\cite{xu2026long,kim2025salova}; models must preserve object identity, retain the latest spatial relations or key events, and determine whether the state persists without contradictory evidence. We define this capability as hidden-state spatiotemporal reasoning: inferring a target's latent state during prolonged invisibility from prior interactions and state changes. However, most video-understanding enhancement methods~\cite{weng2024longvlm,wang2024videoagent,cheng2025enhancing,pang2025mr} remain evidence-driven, aggregating visible observations~\cite{zhang2024simple,you2024toward}, compressing video content~\cite{jiang2025storm,liu2025video}, or retrieving relevant clips~\cite{xu2026long,kim2025salova}, without explicitly modeling disappearance causes or state persistence. Consequently, they often equate ``invisible'' with ``unknown.''

To address this challenge, we propose \textbf{StateTrace}, an object-centric framework for hidden-state reasoning in long videos. For each video, StateTrace performs an offline parsing pass in advance to discover objects, track trajectories, extract spatial relations, and identify state-transition events such as \emph{put-into}, \emph{covered-by}, \emph{occluded-by}, and \emph{removed-from}. These signals are unified into a reusable \emph{spatiotemporal state memory} that captures object appearances, interactions, invisibility causes, and persistent latent states. At inference time, StateTrace retrieves query-relevant evidence from this memory, reconstructs the target state-evolution trajectory, and summarizes it into compact reasoning cues, which are combined with key video segments and global context to support explicit latent-state reasoning.

%Beyond methodology, we argue that existing long-video benchmarks~\cite{rawal2024cinepile,chen2024rextime,cheng2025v,wu2024visual} still substantially under-evaluate this capability. To fill this gap, we introduce \textbf{HSR-Bench}, a new large-scale diagnostic benchmark for \emph{Hidden-State Reasoning} in long videos. HSR-Bench contains 1,427 video-QA samples from 1,384 unique videos and is designed to systematically probe object persistence and latent-state inference under challenging conditions such as occlusion, containment, covering, and long-delay state querying. Extensive experiments demonstrate that StateTrace consistently and substantially improves strong backbone VideoLLMs, including InternVL2.5~\cite{chen2024expanding}, Qwen2.5-VL~\cite{bai2025qwen2}, and VideoLLaMA3~\cite{zhang2025videollama}, across both standard public benchmarks and the newly introduced HSR-Bench. Notably, the gains are especially pronounced on HSR-Bench, where all backbones are improved by approximately 15--28 points, highlighting the particular effectiveness of StateTrace for reasoning over prolonged invisibility and latent object states. Our contributions are summarized as follows:

Beyond methodology, we introduce \textbf{HSR-Bench}, a diagnostic benchmark for \emph{Hidden-State Reasoning} in long videos, since existing benchmarks~\cite{rawal2024cinepile,chen2024rextime,cheng2025v,wu2024visual} under-evaluate this capability. HSR-Bench contains 1,427 video-QA samples from 1,384 videos and targets object persistence and latent-state inference under occlusion, containment, covering, and long-delay querying. Experiments show that StateTrace consistently improves strong VideoLLMs, including InternVL2.5~\cite{chen2024expanding}, Qwen2.5-VL~\cite{bai2025qwen2}, and VideoLLaMA3~\cite{zhang2025videollama}, on both public benchmarks and HSR-Bench, with especially large gains of about 15\%--28\% on HSR-Bench. Our contributions are as follows:

\begin{enumerate}
    \item We propose and systematically formulate the problem of \emph{hidden-state spatiotemporal reasoning} in long-video question answering, identifying persistent state modeling during object invisibility as a core missing capability of existing VideoLLMs and a major source of failure in complex long-horizon spatial reasoning.
    \item We introduce \textbf{StateTrace}, a novel object-centric framework that equips VideoLLMs with an explicit mechanism for hidden-state reasoning. It transforms long-video reasoning from purely direct visual evidence-driven into structured state-centric reasoning over latent object dynamics.
    \item We establish \textbf{HSR-Bench}, a new diagnostic benchmark for hidden-state spatiotemporal reasoning in long videos. HSR-Bench covers diverse challenging scenarios and provides a dedicated testbed for evaluating object persistence and hidden-state inference beyond visible evidence.
\end{enumerate}

\section{Related Work}

\paragraph{\textbf{Vision-Language Models for Video Understanding.}}
%Recent vision-language models for video understanding~\cite{song2024moviechat,chen2024videollm,qian2024streaming,ryoo2024xgen} generally follow a common paradigm: videos are converted into frame-~\cite{shu2025video,li2025videoscan} or clip-level~\cite{li2025improving,li2501videochatflash} visual tokens, which are then aligned with large language models~\cite{brown2020language,chowdhery2023palm,touvron2023llama} for downstream question answering and reasoning. Early works such as \textit{Video-LLaMA}, \textit{Video-ChatGPT}, \textit{Video-LLaVA}, and \textit{VideoChat}~\cite{li2025videochat} established this formulation for video-grounded dialogue and video question answering. Subsequent models further expanded the capability of Video-LLMs. In particular, \textit{LLaVA-Video}~\cite{zhang2024llava} introduced richer video instruction tuning for general-purpose video understanding, while more recent models such as \textit{Qwen2-VL}~\cite{wang2024qwen2}, \textit{VideoLLaMA3}, and \textit{InternVL2.5} improved visual perception and multimodal reasoning through stronger visual encoders, better cross-modal alignment, and more scalable video representations. Despite these advances, existing Video-LLMs mainly excel at general video understanding, but still remain limited in fine-grained spatiotemporal reasoning, especially when target objects become temporarily invisible or heavily occluded.

Recent VLMs for video understanding~\cite{song2024moviechat,chen2024videollm,qian2024streaming,ryoo2024xgen} typically convert videos into frame-~\cite{shu2025video,li2025videoscan} or clip-level~\cite{li2025improving,li2501videochatflash} visual tokens and align them with LLMs~\cite{brown2020language,chowdhery2023palm,touvron2023llama,li2025identify,pmlr-v267-tan25f} for question answering and reasoning. Early systems such as \textit{Video-LLaMA}, \textit{Video-ChatGPT}, \textit{Video-LLaVA}, and \textit{VideoChat}~\cite{li2025videochat} established this paradigm for video-grounded dialogue and video question answering. Later models expanded Video-LLM capabilities: \textit{LLaVA-Video}~\cite{zhang2024llava} introduced richer video instruction tuning, while \textit{Qwen2-VL}~\cite{wang2024qwen2}, \textit{VideoLLaMA3}, and \textit{InternVL2.5} improved perception and multimodal reasoning through stronger visual encoders, tighter cross-modal alignment, and more scalable video representations~\cite{li2023diffnas}. However, these models remain limited in fine-grained spatiotemporal reasoning, especially when target objects are temporarily invisible or heavily occluded.

\paragraph{\textbf{Spatiotemporal-Augmented Vision-Language Models for Video Understanding.}}

To improve spatiotemporal reasoning, recent works augment VLMs along temporal, spatial, and joint dimensions~\cite{li2026vlaattcadaptivetesttimecompute,li2026sentinelvlametacognitivevlamodel}. Temporally, \textit{TimeChat}~\cite{ren2024timechat} introduces timestamp-aware encoding, while \textit{MovieChat}~\cite{song2024moviechat} adopts memory-based designs for long-video understanding. \textit{Video-RAG}~\cite{luo2024video} and \textit{FlexSelect}~\cite{zhang2025flexselect} further enhance long-video reasoning through context retrieval, selection, and compression. Spatially, \textit{PG-Video-LLaVA}~\cite{munasinghe2023pg} provides pixel-level grounding, \textit{VISA}~\cite{yan2024visa} combines language-guided reasoning with mask prediction, and \textit{ViLLa}~\cite{zheng2025villa} models object dynamics with track-level representations. Other studies jointly strengthen spatial and temporal modeling through fine-grained perception and global context~\cite{maaz2024videogpt+}, in-model spatiotemporal dependency modeling~\cite{liu2024st}, or visual supervision with long-context compression~\cite{wang2025internvideo2,li2023stprivacy}. Yet these methods still struggle in occlusion-heavy videos, where reasoning requires persistent hidden-state tracking beyond visible evidence.
\section{Method}
%We propose StateTrace, an object-centric framework for hidden-state spatiotemporal reasoning in videos. Unlike video-segment-centric evidence organization schemes, StateTrace maintains an object-centric spatiotemporal state memory for video reasoning, enabling inference over target objects even when direct visual evidence is interrupted.

StateTrace consists of three stages:

\noindent\textbf{Offline Spatiotemporal State Memory Construction:} This stage parses the video offline to build an object-centric spatiotemporal state memory that stores information like object states, relations, and visibility-transition evidence over time.

\noindent\textbf{Question-Driven State Trajectory Summarization:} Given a question, the system retrieves question-relevant evidence from the state memory and summarizes it into complementary semantic and spatial representations.

\noindent\textbf{Summary-Augmented Answer Generation:} The system combines the retrieved visual evidence and the generated summaries to construct the final input for VideoLLM-based answer generation.

% \noindent\textbf{Offline Spatiotemporal State Memory Construction:} Before question answering, the system performs object-level parsing over the video to construct an object-centric spatiotemporal state memory, storing chunk-level semantic information and relative positional relations, as well as frame-level spatial events and visibility-transition evidence.
%
% \noindent\textbf{Question-Driven State Trajectory Summarization:} Given a question, the system retrieves relevant objects, relations, and event chains from the spatiotemporal state memory, and further produces two complementary summaries: a General Events Summary for cross-segment semantic aggregation and a Spatial Stage Summary for explicit object-state evolution modeling.
%
% \noindent\textbf{Summary-Augmented Answer Generation:} The system constructs a multi-source answer input by combining question-relevant visual evidence with two complementary summary representations, namely a General Events Summary for cross-segment semantic aggregation and a Spatial Stage Summary for explicit object-state evolution modeling, and feeds them into a VideoLLM for final answer generation.

\subsection{Hidden-State Spatiotemporal Reasoning}

Hidden-state spatiotemporal reasoning refers to answering a question \(q\) about a video \(V\) using both visible evidence and the latent state of a temporarily invisible target object. We consider cases where the object is initially visible but later becomes invisible due to occlusion, containment, or covering while remaining in the scene; this period is a \emph{hidden-state interval}. Formally, the task is \(\hat{a}=F(V,q)\), where \(F\) reasons over visible observations and latent object states.
\subsection{Stage I: Offline Spatiotemporal State Memory Construction}

%In implementation, the offline memory is materialized through two reusable graph structures, \((G_V,G_E)\). The first is a chunk-level directed video graph \(G_V\), whose segment nodes correspond to video chunks \(\{C_m\}_{m=1}^{K}\), with \(C_m\) denoting the \(m\)-th chunk. These nodes are the main carriers of memory, storing chunk-level semantic context together with a frame-level spatial timeline, while auxiliary object-linked edges provide a higher-level object-centric view. The second graph, \(G_E\), is an inverted entity index from entity names to chunk sets, enabling efficient access from entity-level queries to the corresponding segments in \(G_V\).

\subsubsection{Graph-Based Organization of the Spatiotemporal Memory}

The reusable offline memory is stored in a chunk-level directed graph, denoted by \(G_V\), whose segment nodes correspond to video chunks \(\{C_m\}_{m=1}^{K}\). Each segment node in \(G_V\) serves as a memory carrier that organizes chunk-level semantic context together with frame-level spatial timelines and event records. In addition, we construct an auxiliary graph \(G_E\), which acts as an entity-level access structure over \(G_V\): it maps entity names to the relevant chunk sets and supports efficient entity-to-segment retrieval, but does not store the memory content.

\begin{figure*}[t]
\centering
\includegraphics[width=0.965\textwidth]{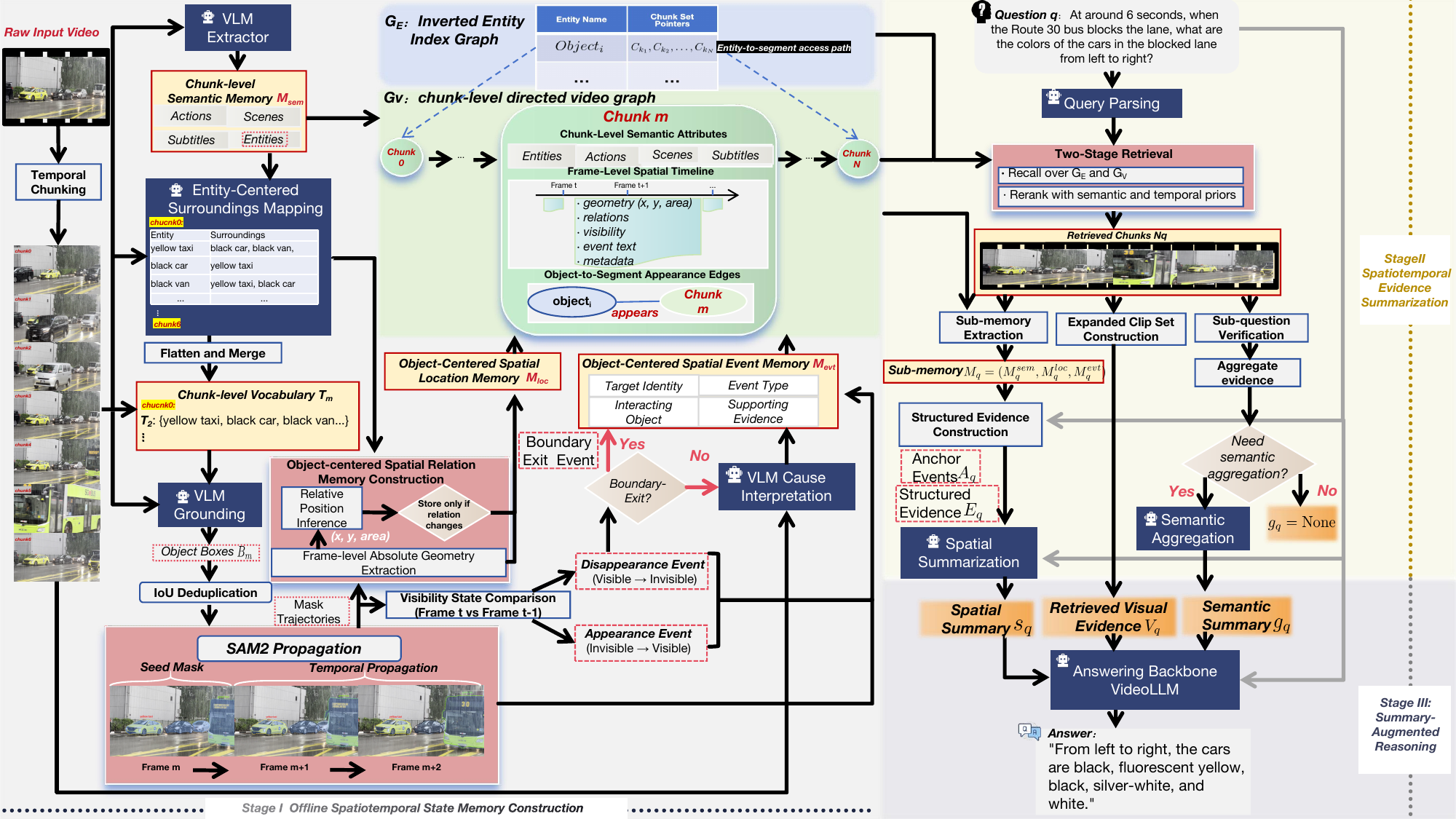}
\vspace{-10pt}
\caption{Overview of StateTrace. The pipeline consists of three stages: (I) offline spatiotemporal state memory construction, (II) query-time retrieval and spatiotemporal evidence summarization, and (III) final answer generation with evidence.}
\Description{The figure illustrates the three-stage StateTrace pipeline. Stage I divides a long video into chunks and constructs an object-centric spatiotemporal memory containing semantic information, object trajectories, spatial relations, visibility changes, and state-transition events. Stage II retrieves question-relevant video chunks and memory records and summarizes them into semantic and spatial evidence. Stage III combines the retrieved visual evidence, the generated summaries, and the question as input to a VideoLLM for final answer generation.}
\vspace{-13pt}
\label{fig:method}
\end{figure*}

For clarity, we decompose the memory into three components:
\begin{equation}
M = (M_{\mathrm{sem}}, M_{\mathrm{loc}}, M_{\mathrm{evt}}),
\label{eq:memory_decomp}
\end{equation}
where \(M_{\mathrm{sem}}\) denotes chunk-level semantic context, \(M_{\mathrm{loc}}\) denotes object-centered spatial location memory, and \(M_{\mathrm{evt}}\) denotes object-centered spatial event memory.

Specifically, \(M_{\mathrm{sem}}\) is stored on the segment nodes of \(G_V\) as semantic attributes, including entities, actions, scenes, and subtitles.

For \(M_{\mathrm{loc}}\), the absolute geometry of object \(o_i\) at sampled frame \(t\) in chunk \(C_m\) is represented as
\begin{equation}
p_i^{m,t} = (x_i^{m,t}, y_i^{m,t}, a_i^{m,t}),
\end{equation}
where \(x_i^{m,t}\) and \(y_i^{m,t}\) are the normalized image-space centroid coordinates of \(o_i\), and \(a_i^{m,t}\) is its area. Pairwise spatial relations are represented as
\begin{equation}
r_{ij}^{m,t} = (o_i, o_j, \rho_{ij}^{m,t}, c_{ij}^{m,t}),
\end{equation}
where \(\rho_{ij}^{m,t}\) denotes the relative spatial relation between \(o_i\) and \(o_j\), and \(c_{ij}^{m,t}\) is an optional confidence score. These location records are written into the frame-level timelines associated with the corresponding segment nodes in \(G_V\), with relations stored only when they change from the previous frame.

For \(M_{\mathrm{evt}}\), the event memory at frame \(t\) in chunk \(C_m\) is represented as
\begin{equation}
e^{m,t} = (v^{m,t}, \epsilon^{m,t}, \mu^{m,t}),
\end{equation}
where \(v^{m,t}\) denotes target visibility, \(\epsilon^{m,t}\) denotes textual event descriptions, and \(\mu^{m,t}\) denotes structured event metadata, including event types, supporting attributes, and evidence such as VLM-inferred disappearance causes. These event records are stored in the same frame-level timelines of the corresponding segment nodes in \(G_V\). In addition, object occurrence is marked through auxiliary object-to-segment \textit{appears} edges in \(G_V\).

In this way, \(G_V\) functions as the main carrier of reusable spatiotemporal memory, while \(G_E\) serves only as an auxiliary access graph that facilitates later entity-centered retrieval and reasoning.

\subsubsection{Object-Centric Spatiotemporal Parsing}

We first divide the video into consecutive chunks \(\{C_m\}_{m=1}^{K}\) and use a VLM extractor \(\mathcal{E}\) to obtain chunk-level semantics:
\begin{equation}
S_m=\mathcal{E}(C_m),
\label{eq:chunk_semantics}
\end{equation}
where \(S_m\), together with aligned subtitles, forms \(M_{\mathrm{sem}}\) for later retrieval and reasoning.

StateTrace then performs object grounding and mask propagation within each chunk, maintaining cross-chunk continuity through IoU-based association. For each semantic entity, we identify surrounding objects and retain the entity-to-surroundings mapping for relative-position reasoning and occlusion-event interpretation. The entities and surrounding objects are flattened into a chunk-level vocabulary \(T_m\) for subsequent grounding.

\paragraph{VLM grounding.}
For each tracking segment, spatial parsing begins from its first frame. Given the chunk-specific grounding vocabulary \(T_m\), we use a VLM grounding module \(\mathcal{G}\) to obtain object-level bounding boxes:
\begin{equation}
B_m=\mathcal{G}(f_{t_m},T_m),
\label{eq:vlm_grounding}
\end{equation}
where \(B_m\) denotes the resulting set of object boxes. Since different lexical items in \(T_m\) may produce highly overlapping detections for the same object, we further apply IoU-based deduplication to remove redundant boxes, yielding the deduplicated box set \(\tilde{B}_m\).

\paragraph{SAM2 propagation.}
We use the SAM2 image predictor \(\mathcal{I}\) to convert the deduplicated boxes on the first frame into seed masks, and then use the SAM2 video predictor \(\mathcal{P}\) to propagate them temporally over the current segment:
\begin{equation}
Z_m^{0}=\mathcal{I}(f_{t_m},\tilde{B}_m),
\label{eq:mask_init}
\end{equation}
\begin{equation}
Z_m=\mathcal{P}(Z_m^{0},C_m),
\label{eq:mask_propagate}
\end{equation}

Cross-chunk continuity is maintained by matching masks propagated from the previous segment with newly generated masks at the current boundary using mask-level IoU. Here, \(Z_m\) denotes the final propagated mask trajectories, which constitute the output of spatiotemporal parsing.

\subsubsection{Object-Centered Spatial Relation Memory Construction}

Based on the obtained spatiotemporal mask trajectories, we construct \(M_{\mathrm{loc}}\) by deriving both absolute object geometry and pairwise spatial relations at the frame level. For each sampled frame in chunk \(C_m\), we first recover the retained target and surrounding objects from the propagated masks, and compute their image-space centroids and areas. Formally, for object \(o_i\) at frame \(t\), its normalized absolute geometry is obtained as
\begin{equation}
x_i^{m,t}=\frac{c_{x,i}^{m,t}}{W^{m,t}},\qquad
y_i^{m,t}=\frac{c_{y,i}^{m,t}}{H^{m,t}},\qquad
a_i^{m,t}=|\Omega_i^{m,t}|,
\label{eq:absolute_geometry_derivation}
\end{equation}
where \(c_{x,i}^{m,t}\) and \(c_{y,i}^{m,t}\) denote the image-space centroid coordinates of \(o_i\), \(W^{m,t}\) and \(H^{m,t}\) denote the frame width and height, and \(|\Omega_i^{m,t}|\) denotes the object area. The target--surroundings mappings are inherited from the object-centric spatiotemporal parsing stage.

Relative positional relations are determined from normalized centroid offsets and inter-object distances. Formally, the positional relation inference can be written as
\begin{equation}
(\rho_{ij}^{m,t}, c_{ij}^{m,t})=
\Phi\!\left(
\Delta x_{ij}^{m,t},
\Delta y_{ij}^{m,t},
d_{ij}^{m,t}
\right),
\label{eq:relative_relation}
\end{equation}

where \(\rho_{ij}^{m,t}\) denotes the discrete relation label between \(o_i\) and \(o_j\), and \((\Delta x_{ij}^{m,t}, \Delta y_{ij}^{m,t}, d_{ij}^{m,t})\) are the corresponding geometric cues. In practice, \(\Phi\) maps them to labels such as \textit{left}, \textit{right}, \textit{above}, or \textit{below}, together with a distance level of \textit{near}, \textit{mid}, or \textit{far}. The continuous geometric quantities are stored as auxiliary fields with the confidence score \(c_{ij}^{m,t}\), which decreases with inter-object distance. To reduce redundancy, only relations whose \((o_i,o_j,\rho_{ij}^{m,t})\) label changes from the previous frame are written into the spatial timeline.

\subsubsection{Object-Centered Spatial Event Memory Construction}

%To construct the spatial event memory \(M_{\mathrm{evt}}\), StateTrace extracts frame-level events from the tracking results and writes them into the event record \(e^{m,t}\). For each target object, we determine its visibility state at each sampled frame and compare it with that in the previous frame. A transition from invisible to visible is treated as an appearance event, whereas a transition from visible to invisible is treated as a disappearance event. These transitions are written into the textual event record \(\epsilon^{m,t}\) and the structured metadata \(\mu^{m,t}\), while the per-target visibility states are recorded in \(v^{m,t}\); object appearance is additionally marked by object-to-segment \textit{appears} edges.

To construct \(M_{\mathrm{evt}}\), StateTrace extracts frame-level events from the tracking results and writes them into \(e^{m,t}\). For each target object, visibility at each sampled frame is compared with the previous frame. Invisible-to-visible transitions are treated as appearance events, and visible-to-invisible transitions as disappearance events. These transitions are written into \(\epsilon^{m,t}\) and \(\mu^{m,t}\), while per-target visibility states are recorded in \(v^{m,t}\); object appearance is additionally marked by object-to-segment \textit{appears} edges.

\paragraph{Boundary-exit filtering.}
For a disappearance event of object \(o_i\), StateTrace first distinguishes boundary exit from within-frame disappearance. Let \(Z_i^{m,t}\) denote the mask of \(o_i\) at its last visible frame, and let \(\partial\Omega\) denote the image boundary region. We compute the boundary contact ratio as
\begin{equation}
\beta_i^{m,t}=
\frac{|Z_i^{m,t}\cap \partial\Omega|}{|Z_i^{m,t}|},
\label{eq:boundary_ratio}
\end{equation}
which measures how much of the object support touches the image boundary. If \(\beta_i^{m,t}\ge\tau_b\), the transition is interpreted as boundary exit and written as a \textit{left-frame} event. Otherwise, StateTrace treats it as a non-boundary disappearance and triggers further cause analysis.

\paragraph{VLM-based disappearance-cause interpretation.}
For each non-boundary disappearance, StateTrace infers its cause from a short video-context centered on the disappearance moment and assembled from a configurable number of preceding and following chunks:
\begin{equation}
\hat{\tau}_{i,\mathrm{disp}}^{m,t}
=
D(\{f_{t-k},\ldots,f_t\},o_i,\mathrm{context}),
\label{eq:disappearance_cause}
\end{equation}
where \(D\) is the disappearance-cause inference module. The predicted cause is mapped to one of four outcomes: \textit{inside}, \textit{occluded}, \textit{other}, or \textit{unknown}. For \textit{inside} and \textit{occluded}, the interacting object can be further aligned to candidates derived from the target--surroundings mapping. The final result is written into \(\epsilon^{m,t}\) and \(\mu^{m,t}\), including the event type, target identity, interacting object when available, and supporting evidence.

\subsection{Stage II: Question-Driven Spatiotemporal Evidence Summarization}

Instead of directly using the full spatiotemporal state memory \(M\), we retrieve a question-relevant subset and construct a summary representation \(S_q\). This stage produces two complementary summaries: a semantic summary, when available, that condenses answer-relevant information from retrieved segments, and a spatial summary that captures spatiotemporal evidence from selected video segments. Together, they support answer generation while reducing key-information selection difficulty in long contexts.

\subsubsection{Question-Guided Retrieval and Evidence Extraction}

Given a question \(q\), StateTrace converts it into a structured query:
\begin{equation}
\Psi(q) = (Q_q, \ell_q, t_q),
\label{eq:structured_query}
\end{equation}
where \(Q_q\) contains semantic query items, \(\ell_q\) contains retrieval and reasoning controls, and \(t_q\) specifies temporal constraints when available. We obtain \(\Psi(q)\) by prompting a language model to extract keywords, control signals, and temporal specifications, followed by rule-based normalization.

StateTrace then retrieves relevant chunks in two stages. It first forms an initial candidate set by combining entity-to-chunk lookup through \(G_E\) with semantic matching over segment attributes in \(G_V\):
\begin{equation}
N_q^{(0)} = R_{\mathrm{recall}}(Q_q,\ell_q,t_q;G_V,G_E).
\label{eq:recall_stage}
\end{equation}
The candidates \(N_q^{(0)}\) are then reranked by semantic similarity. When temporal constraints are available, \(t_q\) serves as a soft prior favoring chunks consistent with the specified time or coarse anchors near the video beginning or end:
\begin{equation}
N_q = R_{\mathrm{rank}}(N_q^{(0)},Q_q,t_q;G_V).
\label{eq:rank_stage}
\end{equation}
The resulting set \(N_q\) is used to extract the corresponding fields from \(G_V\), forming the question-relevant sub-memory:
\begin{equation}
M_q = (M_{\mathrm{sem}}^{q}, M_{\mathrm{loc}}^{q}, M_{\mathrm{evt}}^{q}),
\label{eq:sub_memory}
\end{equation}
where \(M_{\mathrm{sem}}^{q}\), \(M_{\mathrm{loc}}^{q}\), and \(M_{\mathrm{evt}}^{q}\) are the semantic attributes, spatial timelines, and event records stored on the retrieved segment nodes. Thus, retrieval operates over graph-organized memory, while evidence extraction selects the relevant fields associated with the retrieved chunks.

\subsubsection{Semantic Aggregation and Spatial Summarization}

StateTrace compresses the retrieved sub-memory \(M_q\) into two complementary question-driven summaries for the reasoning model: a semantic aggregation summary \(g_q\) and a spatial summary \(s_q\).

For semantic aggregation, StateTrace verifies decomposed sub-questions over the retrieved chunks and organizes answer-relevant evidence into an aggregated text \(x_q^{\mathrm{agg}}\). When multi-segment reasoning is required and \(x_q^{\mathrm{agg}}\) is non-empty, a language model compresses it into:
\begin{equation}
g_q = \Gamma(x_q^{\mathrm{agg}}, q),
\label{eq:semantic_aggregation}
\end{equation}
where \(\Gamma\) maps the aggregated evidence and question to the final semantic summary.

For spatial summarization, StateTrace expands the top retrieved chunks with disappearance-related chunks and their immediate predecessors, then merges their frame-level timelines. From these records, it constructs structured spatial--temporal evidence \(E_q\), which compacts salient relations, repairs visibility states, and preserves frame-level object, event, and visibility information. It also extracts anchor events \(A_q\), a sparse set of key disappearance-, occlusion-, entry-, and reappearance-related transitions with corresponding frames. Thus, \(E_q\) provides structured temporal context, while \(A_q\) highlights critical state transitions. The spatial summary is generated as:
\begin{equation}
s_q = S(E_q, A_q, q),
\label{eq:spatial_summary}
\end{equation}
where \(S\) denotes spatial summarization. In implementation, \(S\) prompts a multimodal model with \(E_q\), \(A_q\), allowed frame indices, and the corresponding video clip, followed by post-processing for summary formatting, visibility repair, strong-claim sanitization, and frame-range filtering.

\subsection{Stage III: Summary-Augmented Answer}

The third stage of StateTrace performs final answer generation by combining the summaries from Stage II with the retrieved visual evidence. Given a question \(q\), aligned subtitle context \(S_q^{\mathrm{sub}}\), and retrieved visual evidence \(V_q\), StateTrace constructs a multimodal answer input. Its textual part integrates the question, subtitle context, and the summaries \((g_q, s_q)\), while its visual part consists of the corresponding video chunks.

The final prediction is then generated by a VideoLLM \(M_{\mathrm{ans}}\) as
\begin{equation}
\hat{a} = M_{\mathrm{ans}}(q, S_q^{\mathrm{sub}}, V_q, g_q, s_q),
\label{eq:final_answer}
\end{equation}
This formulation is model-agnostic and can be instantiated with different VideoLLM backbones, as examined in the experiments.
\section{HSR-Benchmark Construction}

%We construct HSR-Bench as a diagnostic benchmark for hidden-state spatiotemporal reasoning under occlusion and invisibility. It contains 1,427 video-QA samples from 1,384 unique videos, and each sample consists of a video, a natural-language question, and four candidate answers. Most samples involve substantial visibility interruption, with 52.35\% heavy occlusion, 46.11\% medium occlusion, and only 1.54\% light occlusion, making the benchmark challenging for models that rely mainly on direct visual evidence. Rather than organizing the benchmark by semantic topic, we define four reasoning-oriented tasks: occluded entity recognition, occlusion event summary, occlusion-conditioned attribute extraction, and post-occlusion state persistence. HSR-Bench is built from two public video sources, OVIS~\cite{qi2022occluded} and MOSEv2~\cite{ding2025mosev2}, which are allocated to different tasks according to their annotation characteristics. After standardizing all videos into a unified representation, we mine task-specific candidate instances using annotation-driven, rule-based procedures over masks, visibility statistics, and temporal constraints, without relying on a vision-language model. We then use Qwen3-VL~\cite{bai2025qwen3} to draft task-specific questions, followed by manual answer annotation and distractor construction. Detailed construction procedures are provided in \textbf{Appendix}.

We construct HSR-Bench as a diagnostic benchmark for hidden-state spatiotemporal reasoning under occlusion and invisibility. It contains 1,427 video-QA samples from 1,384 unique videos, each with a video, a natural-language question, and four candidate answers. Most samples involve substantial visibility interruption, including 52.35\% heavy and 46.11\% medium occlusion. HSR-Bench covers four tasks: occluded entity recognition, occlusion event summary, occlusion-conditioned attribute extraction, and post-occlusion state persistence. Built from OVIS~\cite{qi2022occluded} and MOSEv2~\cite{ding2025mosev2}, it uses rule-based mining over masks, visibility statistics, and temporal constraints, followed by Qwen3-VL~\cite{bai2025qwen3}-assisted question drafting, manual answer annotation, and distractor construction. Details are provided in \textbf{Appendix}.

\begin{table}[t]
\vspace{-8pt}
\caption{Main results on public video QA benchmarks. StateTrace is compared with base VideoLLMs and other enhancement methods across different benchmarks. The best result in each column is shown in bold.}
\vspace{-6pt}
\label{tab:main_results}
\small
\centering
\setlength{\tabcolsep}{3pt}
\renewcommand{\arraystretch}{0.86}
\begin{tabularx}{\columnwidth}{@{}Xccccc@{}}
\toprule
Model & Size & MLVU & \multicolumn{2}{c}{VideoMME} & LVB \\
\cmidrule(lr){4-5}
& & & w/o sub. & w/ sub. & \\
\midrule
InternVL2.5~\cite{chen2024expanding} & 2B & 61.4 & 51.9 & 54.1 & 52.0 \\
InternVL2.5 + Video-RAG~\cite{luo2024video} & 2B & 62.4 & 52.4 & 54.8 & 53.1 \\
InternVL2.5 + FlexSelect~\cite{zhang2025flexselect} & 2B & 63.0 & 52.6 & 55.3 & 54.0 \\
\rowcolor{gray!12}
InternVL2.5 + StateTrace \textbf{(Ours)} & 2B & 64.0 & 52.8 & 56.0 & 56.3 \\
\midrule

VideoLLaMA3~\cite{zhang2025videollama} & 2B & 65.4 & 59.6 & 63.4 & 57.1 \\
VideoLLaMA3 + Video-RAG~\cite{luo2024video} & 2B & 66.5 & 60.0 & 64.2 & 57.9 \\
VideoLLaMA3 + FlexSelect~\cite{zhang2025flexselect} & 2B & 67.4 & 60.2 & 64.8 & 58.8 \\
\rowcolor{gray!12}
VideoLLaMA3 + StateTrace \textbf{(Ours)} & 2B & 68.9 & 60.5 & 65.4 & 61.0 \\
\midrule

Qwen2.5-VL~\cite{bai2025qwen2} & 3B & 68.2 & 61.5 & 67.6 & 54.2 \\
Qwen2.5-VL + Video-RAG~\cite{luo2024video} & 3B & 69.3 & 61.8 & 68.2 & 55.4 \\
Qwen2.5-VL + FlexSelect~\cite{zhang2025flexselect} & 3B & 70.1 & 62.0 & 68.7 & 56.3 \\
\rowcolor{gray!12}
Qwen2.5-VL + StateTrace \textbf{(Ours)} & 3B & 71.8 & 62.3 & 69.2 & 59.8 \\
\midrule

VideoLLaMA3~\cite{zhang2025videollama} & 7B & 73.0 & 66.2 & 70.3 & 59.8 \\
VideoLLaMA3 + Video-RAG~\cite{luo2024video} & 7B & 74.1 & 67.0 & 71.2 & 60.8 \\
VideoLLaMA3 + FlexSelect~\cite{zhang2025flexselect} & 7B & 75.3 & 68.1 & 72.4 & 62.2 \\
\rowcolor{gray!12}
VideoLLaMA3 + StateTrace \textbf{(Ours)} & 7B & \textbf{77.2} & \textbf{69.7} & \textbf{74.1} & \textbf{64.5} \\
\midrule

Qwen2.5-VL~\cite{bai2025qwen2} & 7B & 68.8 & 65.1 & 71.1 & 56.0 \\
Qwen2.5-VL + Video-RAG~\cite{luo2024video} & 7B & 70.5 & 65.6 & 71.9 & 57.6 \\
Qwen2.5-VL + FlexSelect~\cite{zhang2025flexselect} & 7B & 72.5 & 65.8 & 72.6 & 62.4 \\
\rowcolor{gray!12}
Qwen2.5-VL + StateTrace \textbf{(Ours)} & 7B & 75.6 & 66.0 & 73.0 & 62.8 \\
\midrule

LLaVA-Video~\cite{zhang2024llava} & 7B & 70.8 & 63.3 & 69.7 & 58.2 \\
LLaVA-Video + Video-RAG~\cite{luo2024video} & 7B & 72.4 & 64.5 & 71.0 & 58.7 \\
LLaVA-Video + FlexSelect~\cite{zhang2025flexselect} & 7B & 73.2 & 65.0 & 68.9 & 61.9 \\
\rowcolor{gray!12}
LLaVA-Video + StateTrace \textbf{(Ours)} & 7B & 74.6 & 65.4 & 72.3 & 62.1 \\
\midrule

InternVL2.5~\cite{chen2024expanding} & 8B & 68.9 & 64.2 & 66.9 & 60.0 \\
InternVL2.5 + Video-RAG~\cite{luo2024video} & 8B & 70.3 & 64.8 & 67.7 & 61.2 \\
InternVL2.5 + FlexSelect~\cite{zhang2025flexselect} & 8B & 71.9 & 65.3 & 68.9 & 60.1 \\
\rowcolor{gray!12}
InternVL2.5 + StateTrace \textbf{(Ours)} & 8B & 73.4 & 65.6 & 70.0 & 64.0 \\
\bottomrule
\end{tabularx}
\vspace{-10pt}
\end{table}

\section{Experiments}
\subsection{Evaluation Data and Models.}
We evaluate StateTrace on public video QA benchmarks, including MLVU~\cite{zhou2025mlvu}, VideoMME~\cite{fu2025video} with and without subtitles, and LongVideoBench~\cite{wu2024longvideobench}, as well as HSR-Bench for hidden-state reasoning diagnostics. We instantiate StateTrace with InternVL2.5~\cite{chen2024expanding}, VideoLLaMA3~\cite{zhang2025videollama}, Qwen2.5-VL~\cite{bai2025qwen2}, and LLaVA-Video~\cite{zhang2024llava}, and compare it with Video-RAG~\cite{luo2024video} and FlexSelect~\cite{zhang2025flexselect} where applicable. %Unless otherwise specified, all compared methods share the same answer format, evaluation protocol, and inference setting for fair comparison.

\subsection{Evaluation Metrics}
%For QA performance, we report overall accuracy on all benchmarks. In addition, we conduct two task-specific fine-grained analyses: length-bucket analysis on LongVideoBench to evaluate robustness under longer temporal contexts, and occlusion-severity-bucket analysis on HSR-Bench to evaluate robustness under different levels of occlusion. For efficiency evaluation, we report the total runtime per video and the amortized runtime per question in multi-query settings. All runtime results are measured on the same hardware platform using NVIDIA A100 80GB GPUs under identical inference settings.
For QA performance, we report overall accuracy on all benchmarks. We further conduct length-bucket analysis on LongVideoBench and occlusion-severity-bucket analysis on HSR-Bench.

\begin{table}[t]
\vspace{-8pt}
\caption{Performance on HSR-Bench. StateTrace is compared with base VideoLLMs and representative enhancement methods. The best result is highlighted in bold.}
\vspace{-6pt}
\label{tab:hsr_bench_main}
\small
\centering
\setlength{\tabcolsep}{3pt}
\renewcommand{\arraystretch}{0.85}
\begin{tabularx}{\columnwidth}{@{}Xcc@{}}
\toprule
Model & Size & HSR-Bench \\
\midrule
InternVL2.5~\cite{chen2024expanding} & 2B & 20.81 \\
InternVL2.5 + Video-RAG~\cite{luo2024video} & 2B & 23.46 \\
InternVL2.5 + FlexSelect~\cite{zhang2025flexselect} & 2B & 26.18 \\
\rowcolor{gray!12}
InternVL2.5 + StateTrace \textbf{(Ours)} & 2B & 49.12 \\
\midrule

VideoLLaMA3~\cite{zhang2025videollama} & 2B & 24.73 \\
VideoLLaMA3 + Video-RAG~\cite{luo2024video} & 2B & 27.95 \\
VideoLLaMA3 + FlexSelect~\cite{zhang2025flexselect} & 2B & 31.42 \\
\rowcolor{gray!12}
VideoLLaMA3 + StateTrace \textbf{(Ours)} & 2B & 53.08 \\
\midrule

Qwen2.5-VL~\cite{bai2025qwen2} & 3B & 29.85 \\
Qwen2.5-VL + Video-RAG~\cite{luo2024video} & 3B & 33.11 \\
Qwen2.5-VL + FlexSelect~\cite{zhang2025flexselect} & 3B & 36.84 \\
\rowcolor{gray!12}
Qwen2.5-VL + StateTrace \textbf{(Ours)} & 3B & 57.74 \\
\midrule

VideoLLaMA3~\cite{zhang2025videollama} & 7B & 39.59 \\
VideoLLaMA3 + Video-RAG~\cite{luo2024video} & 7B & 42.37 \\
VideoLLaMA3 + FlexSelect~\cite{zhang2025flexselect} & 7B & 46.21 \\
\rowcolor{gray!12}
VideoLLaMA3 + StateTrace \textbf{(Ours)} & 7B & \textbf{64.19} \\
\midrule

Qwen2.5-VL~\cite{bai2025qwen2} & 7B & 35.88 \\
Qwen2.5-VL + Video-RAG~\cite{luo2024video} & 7B & 38.20 \\
Qwen2.5-VL + FlexSelect~\cite{zhang2025flexselect} & 7B & 41.50 \\
\rowcolor{gray!12}
Qwen2.5-VL + StateTrace \textbf{(Ours)} & 7B & 61.60 \\
\midrule

LLaVA-Video~\cite{zhang2024llava} & 7B & 37.10 \\
LLaVA-Video + Video-RAG~\cite{luo2024video} & 7B & 38.94 \\
LLaVA-Video + FlexSelect~\cite{zhang2025flexselect} & 7B & 43.68 \\
\rowcolor{gray!12}
LLaVA-Video + StateTrace \textbf{(Ours)} & 7B & 60.92 \\
\midrule

InternVL2.5~\cite{chen2024expanding} & 8B & 37.42 \\
InternVL2.5 + Video-RAG~\cite{luo2024video} & 8B & 39.50 \\
InternVL2.5 + FlexSelect~\cite{zhang2025flexselect} & 8B & 43.00 \\
\rowcolor{gray!12}
InternVL2.5 + StateTrace \textbf{(Ours)} & 8B & 52.56 \\
\bottomrule
\end{tabularx}
\vspace{-10pt}
\end{table}

\subsection{Main Results on Public Benchmarks and HSR-Bench}

Table~\ref{tab:main_results} shows that StateTrace consistently improves the corresponding base VideoLLMs across all evaluated backbones on public benchmarks, and the gains are especially clear on LongVideoBench. This pattern suggests that StateTrace is particularly helpful when answering long-video questions that require maintaining state continuity over extended temporal spans. A representative example is VideoLLaMA3-7B, where StateTrace improves LongVideoBench performance from 59.8 to 64.5, while also delivering the best overall results in Table~\ref{tab:main_results}. Similar improvements are also observed on MLVU and VideoMME, indicating that the proposed framework remains effective beyond a single benchmark or model family.

Table~\ref{tab:hsr_bench_main} further shows that the advantage of StateTrace becomes much more substantial on HSR-Bench, where hidden-state reasoning under occlusion is the primary challenge. Across all tested backbones, StateTrace improves over the corresponding base models by 15.14 to 28.31 points. The largest gain is obtained on Qwen2.5-VL-3B, which improves from 29.85 to 57.74, while VideoLLaMA3-7B with StateTrace achieves the best overall result of 64.19. Compared with representative enhancement methods, the margin is also markedly larger on HSR-Bench; for example, on Qwen2.5-VL-7B, Video-RAG and FlexSelect reach 38.20 and 41.50, respectively, both far below StateTrace at 61.60. These results indicate that explicit state memory and occlusion-aware event modeling are particularly important for hidden-state spatiotemporal reasoning.

\begin{figure}[t]
\centering
\includegraphics[width=\columnwidth]{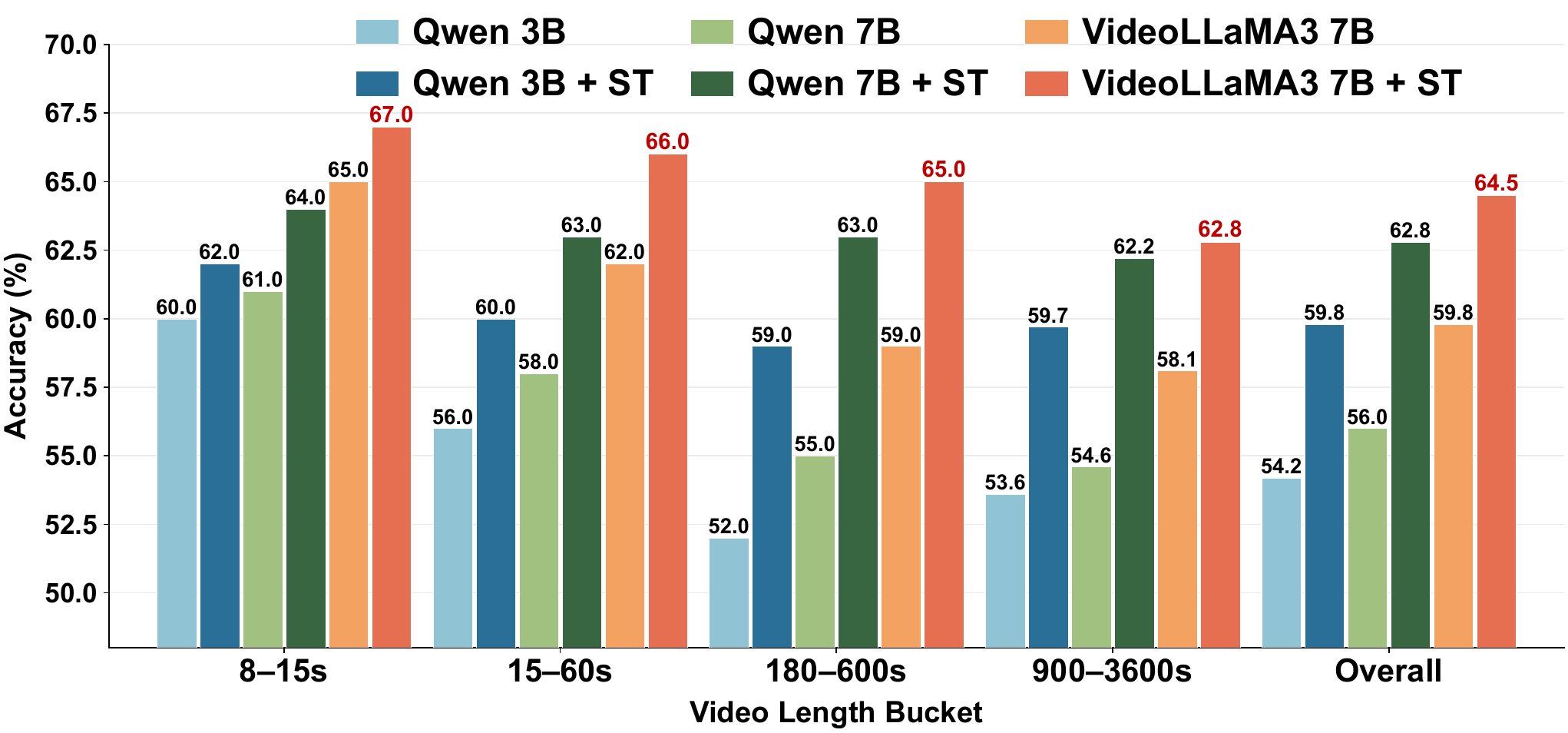}
\vspace{-21pt}
\caption{Performance across video-length buckets on LongVideoBench. ST denotes the proposed StateTrace framework, and Overall is the weighted average accuracy across all duration buckets.}
\Description{The figure compares the accuracy of baseline VideoLLMs and their StateTrace-enhanced variants across multiple video-duration buckets on LongVideoBench. The final group reports the weighted average accuracy over all duration buckets.}
\vspace{-9pt}
\label{fig:lvb_length_main}
\end{figure}

\subsection{Length-Bucket Analysis on LongVideoBench}

To further analyze robustness under longer temporal contexts, we group LongVideoBench samples into its four official duration ranges and visualize representative results in Figure~\ref{fig:lvb_length_main}.

As shown in Figure~\ref{fig:lvb_length_main}, StateTrace consistently improves over the corresponding baselines across all length buckets, with especially clear gains as video duration increases. The effect is most evident in the longest bucket, where long-range temporal reasoning is most challenging: VideoLLaMA3-7B improves from 58.1 to 62.8, and Qwen2.5-VL-7B improves from 54.6 to 62.2. Consistent improvements are also observed in the shorter buckets. These results suggest that StateTrace is particularly beneficial in extended temporal contexts, where its offline graph preserves cross-segment entities, relations, and event evidence, and its question-driven retrieval and summarization help concentrate answer-relevant cues.

\begin{figure}[t]
\centering
\includegraphics[width=\columnwidth]{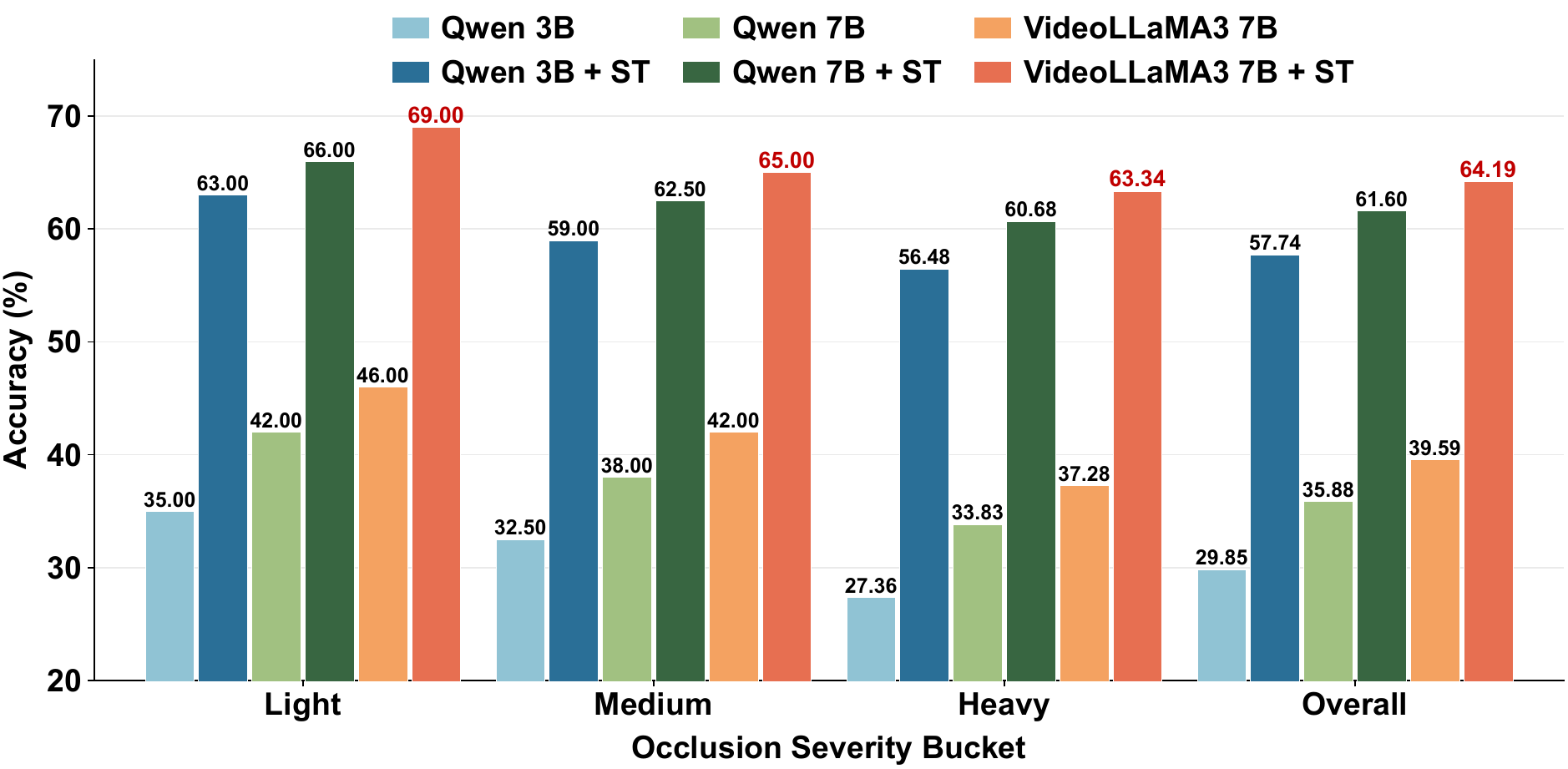}
\vspace{-22pt}
\caption{Performance across occlusion-severity buckets on HSR-Bench. ST denotes StateTrace, and Overall is the weighted average accuracy over all samples.}
\Description{The figure compares the accuracy of baseline VideoLLMs and their StateTrace-enhanced variants across light, medium, and heavy occlusion severity buckets on HSR-Bench. The final group reports the weighted average accuracy over all samples.}
\vspace{-12pt}
\label{fig:hsr_severity}
\end{figure}

% \subsection{Occlusion-Severity-Bucket Analysis on HSR-Bench}

% To further analyze where the gains on HSR-Bench mainly come from, we group evaluation samples into three occlusion-severity levels: light, medium, and heavy. The corresponding results are shown in Table~\ref{tab:hsr_severity}. Since the light bucket contains only 22 samples, our main analysis focuses on the medium and heavy buckets.

% As shown in Table~\ref{tab:hsr_severity}, StateTrace consistently outperforms the corresponding baselines across all occlusion-severity levels. Its advantage is particularly evident under heavy occlusion, where hidden-state reasoning becomes most difficult. For example, under heavy occlusion, VideoLLaMA3-7B improves from 37.3 to 63.4, and Qwen2.5-VL-7B improves from 33.8 to 60.7 after adding StateTrace. Similar gains are also observed in the medium bucket, where VideoLLaMA3-7B improves from 42.0 to 65.0 and Qwen2.5-VL-7B from 38.0 to 62.5. This trend suggests that the advantage of StateTrace comes from its object-centric spatiotemporal state memory, which makes hidden object states more recoverable when visibility is severely interrupted by occlusion.
\subsection{Occlusion-Severity-Bucket Analysis on HSR-Bench}

For a more fine-grained analysis, we partition HSR-Bench into three occlusion-severity buckets---light, medium, and heavy---and report the results in Figure~\ref{fig:hsr_severity}.

Figure~\ref{fig:hsr_severity} shows a clear and consistent pattern: StateTrace improves performance across all severity levels, with the largest gains appearing under heavy occlusion, where hidden-state reasoning is most critical. A representative example is VideoLLaMA3-7B, whose accuracy rises from 37.28 to 63.34 in the heavy bucket. Similar tendency is observed for the other backbones. This trend suggests that the advantage of StateTrace comes from its object-centric spatiotemporal state memory, which makes hidden object states more recoverable when visibility is severely interrupted by occlusion.

\subsection{Ablation Study}
We ablate the key components of the proposed pipeline. Unless otherwise noted, all settings are kept the same as the full model except for the modified component. Overall, each component contributes positively, with the full pipeline showing the most consistent gains on LongVideoBench.
\begin{table}[t]
\vspace{-4pt}
\caption{Ablation on disappearance-cause reasoning under different backbones. DispC denotes disappearance-cause reasoning. The best result in each metric is shown in bold.}
\vspace{-12pt}
\label{tab:ablation_reason}
\small
\centering
\setlength{\tabcolsep}{3pt}
\renewcommand{\arraystretch}{0.85}
\begin{tabularx}{\columnwidth}{@{}Xcccccc@{}}
\toprule
Backbone & Size & Setting & MLVU & \multicolumn{2}{c}{VideoMME} & LVB \\
\cmidrule(lr){5-6}
& & & & w/o sub. & w/ sub. & \\
\midrule
InternVL2.5~\cite{chen2024expanding} & 2B & w/o DispC & 62.3 & 52.1 & 55.0 & 53.7 \\
\rowcolor{gray!12}
InternVL2.5 & 2B & Full & 64.0 & 52.8 & 56.0 & 56.3 \\
\midrule
InternVL2.5 & 8B & w/o DispC & 71.4 & 64.6 & 68.7 & 61.3 \\
\rowcolor{gray!12}
InternVL2.5 & 8B & Full & 73.4 & 65.6 & 70.0 & 64.0 \\
\midrule
Qwen2.5-VL~\cite{bai2025qwen2} & 3B & w/o DispC & 70.0 & 61.5 & 68.1 & 56.9 \\
\rowcolor{gray!12}
Qwen2.5-VL & 3B & Full & 71.8 & 62.3 & 69.2 & 59.8 \\
\midrule
Qwen2.5-VL & 7B & w/o DispC & 73.5 & 65.0 & 71.6 & 59.7 \\
\rowcolor{gray!12}
Qwen2.5-VL & 7B & Full & 75.6 & 66.0 & 73.0 & 62.8 \\
\midrule
VideoLLaMA3~\cite{zhang2025videollama} & 7B & w/o DispC & 74.8 & 67.6 & 72.5 & 61.2 \\
\rowcolor{gray!12}
VideoLLaMA3 & 7B & Full & \textbf{77.2} & \textbf{69.7} & \textbf{74.1} & \textbf{64.5} \\
\bottomrule
\end{tabularx}
\vspace{-8pt}
\end{table}

\begin{table}[t]
\vspace{-1pt}
\caption{Ablation on spatial summary generation under different backbones. SpS denotes spatial summary. The best result in each metric column is shown in bold.}
\vspace{-10pt}
\label{tab:ablation_summary}
\small
\centering
\setlength{\tabcolsep}{3pt}
\renewcommand{\arraystretch}{0.85}
\begin{tabularx}{\columnwidth}{@{}Xcccccc@{}}
\toprule
Backbone & Size & Setting & MLVU & \multicolumn{2}{c}{VideoMME} & LVB \\
\cmidrule(lr){5-6}
& & & & w/o sub. & w/ sub. & \\
\midrule
InternVL2.5~\cite{chen2024expanding} & 2B & w/o SpS & 61.8 & 51.8 & 54.4 & 52.9 \\
\rowcolor{gray!12}
InternVL2.5 & 2B & Full & 64.0 & 52.8 & 56.0 & 56.3 \\
\midrule
InternVL2.5 & 8B & w/o SpS & 70.8 & 64.2 & 68.2 & 60.8 \\
\rowcolor{gray!12}
InternVL2.5 & 8B & Full & 73.4 & 65.6 & 70.0 & 64.0 \\
\midrule
Qwen2.5-VL~\cite{bai2025qwen2} & 3B & w/o SpS & 69.5 & 61.1 & 67.8 & 56.1 \\
\rowcolor{gray!12}
Qwen2.5-VL & 3B & Full & 71.8 & 62.3 & 69.2 & 59.8 \\
\midrule
Qwen2.5-VL & 7B & w/o SpS & 72.9 & 64.6 & 70.9 & 59.0 \\
\rowcolor{gray!12}
Qwen2.5-VL & 7B & Full & 75.6 & 66.0 & 73.0 & 62.8 \\
\midrule
VideoLLaMA3~\cite{zhang2025videollama} & 7B & w/o SpS & 74.1 & 67.1 & 72.0 & 60.5 \\
\rowcolor{gray!12}
VideoLLaMA3 & 7B & Full & \textbf{77.2} & \textbf{69.7} & \textbf{74.1} & \textbf{64.5} \\
\bottomrule
\end{tabularx}
\vspace{-12pt}
\end{table}
%\subsubsection{Ablation of Disappearance-Cause Reasoning}

% This ablation removes the explicit interpretation of disappearance causes during visibility-to-invisibility transitions. Consequently, the system no longer differentiates between disappearance events induced by leaving the frame, occlusion, entering another object, or other causes, and instead retains only coarse-grained appearance and disappearance events.

% Table~\ref{tab:ablation_reason} shows that removing disappearance-cause reasoning consistently degrades performance across all tested backbones and benchmarks. The most pronounced degradation appears on LongVideoBench, where the drop reaches 3.4 points for VideoLLaMA3-7B (64.5 to 61.1) and 2.9 points for Qwen2.5-VL-7B (62.8 to 59.9). Notably, the effect remains substantial even for the strongest models, indicating that disappearance-cause reasoning provides information that cannot be reliably inferred from coarse visibility-transition evidence alone. A similar but smaller trend is observed on MLVU, where VideoLLaMA3-7B decreases from 77.2 to 75.3 and Qwen2.5-VL-7B from 75.6 to 73.9. These results indicate that knowing that an object disappears is not sufficient; reasoning about why it becomes invisible is crucial for recovering latent state relations over time.

\subsubsection{Ablation of Disappearance-Cause Reasoning}
This ablation removes explicit disappearance-cause reasoning during visibility-to-invisibility transitions, retaining only coarse appearance and disappearance events.

\begin{figure*}[t]
    \centering
    \includegraphics[width=\textwidth]{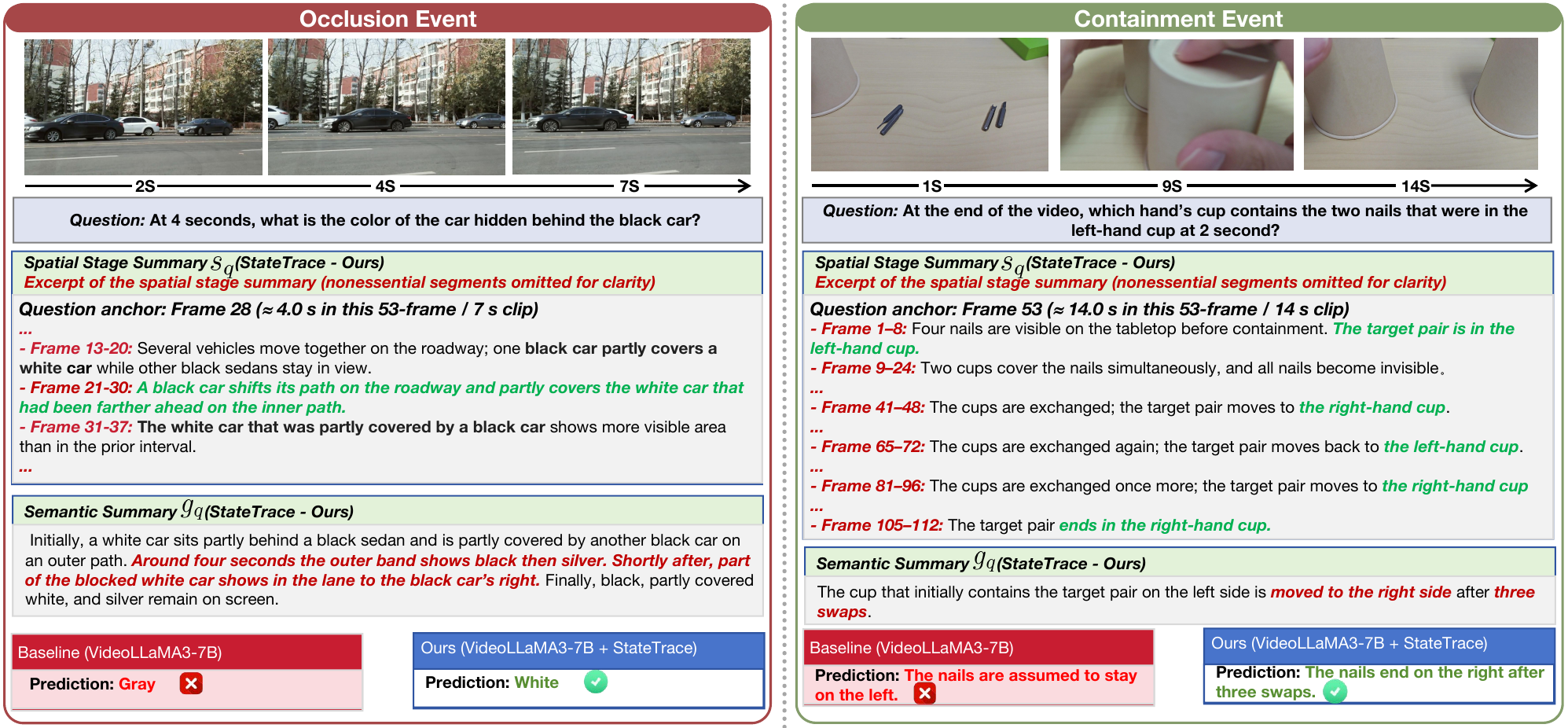}
    \vspace{-22pt}
    \caption{
Two examples illustrating improved hidden-state spatiotemporal reasoning with StateTrace.
    }
    \Description{The figure presents two qualitative examples comparing a baseline VideoLLaMA3-7B model with the StateTrace-enhanced version. The first example involves occlusion, where the model must infer the color of a car after it becomes temporarily hidden behind a black car. The second example involves containment, where the model must determine which cup contains the target nails after repeated exchanges. In both cases, the baseline model gives incorrect answers, while StateTrace preserves hidden object states and answers correctly.}
    \vspace{-10pt}
    \label{fig:case_study}
\end{figure*}
\begin{table}[t]

\caption{Ablation on key clip retrieval under different backbones. KCR denotes key clip retrieval. The best result in each metric column is shown in bold.}
\vspace{-10pt}
\label{tab:ablation_retrieval}
\small
\centering
\setlength{\tabcolsep}{3pt}
\renewcommand{\arraystretch}{0.8}
\begin{tabularx}{\columnwidth}{@{}Xcccccc@{}}
\toprule
Backbone & Size & Setting & MLVU & \multicolumn{2}{c}{VideoMME} & LVB \\
\cmidrule(lr){5-6}
& & & & w/o sub. & w/ sub. & \\
\midrule
InternVL2.5~\cite{chen2024expanding} & 2B & w/o KCR & 62.8 & 52.3 & 55.3 & 54.8 \\
\rowcolor{gray!12}
InternVL2.5 & 2B & Full & 64.0 & 52.8 & 56.0 & 56.3 \\
\midrule
InternVL2.5 & 8B & w/o KCR & 72.1 & 65.0 & 68.9 & 62.4 \\
\rowcolor{gray!12}
InternVL2.5 & 8B & Full & 73.4 & 65.6 & 70.0 & 64.0 \\
\midrule
Qwen2.5-VL~\cite{bai2025qwen2} & 3B & w/o KCR & 70.7 & 61.9 & 68.5 & 58.2 \\
\rowcolor{gray!12}
Qwen2.5-VL & 3B & Full & 71.8 & 62.3 & 69.2 & 59.8 \\
\midrule
Qwen2.5-VL & 7B & w/o KCR & 74.2 & 65.5 & 72.0 & 61.2 \\
\rowcolor{gray!12}
Qwen2.5-VL & 7B & Full & 75.6 & 66.0 & 73.0 & 62.8 \\
\midrule
VideoLLaMA3~\cite{zhang2025videollama} & 7B & w/o KCR & 75.8 & 68.1 & 72.9 & 62.7 \\
\rowcolor{gray!12}
VideoLLaMA3 & 7B & Full & \textbf{77.2} & \textbf{69.7} & \textbf{74.1} & \textbf{64.5} \\
\bottomrule
\end{tabularx}
\vspace{-12pt}
\end{table}

Table~\ref{tab:ablation_reason} shows that removing this component consistently degrades performance across all tested backbones and benchmarks. The effect is most pronounced on LongVideoBench. For example, VideoLLaMA3-7B drops from 64.5 to 61.2, while Qwen2.5-VL-7B drops from 62.8 to 59.7. MLVU and VideoMME also show consistent declines after removing disappearance-cause reasoning. These results indicate that coarse visibility transitions alone do not provide sufficient support for hidden-state reasoning, and that explicitly modeling the cause of invisibility helps recover object states more reliably over time.

% \subsubsection{Ablation of Spatial Summary Generation}

% This ablation removes the Spatial Stage Summary generation module. Instead of compressing structured spatial timeline evidence into a compact summary representation, the model directly consumes the structured spatial evidence.

% Table~\ref{tab:ablation_summary} shows that removing spatial summary generation leads to a consistent but comparatively modest performance drop across all tested backbones and benchmarks. The degradation is generally small but stable, indicating that structured spatial evidence is already useful even without explicit summarization. The clearest drop appears on LongVideoBench, where Qwen2.5-VL-3B decreases from 59.8 to 58.7 and InternVL2.5-2B from 56.3 to 55.4. This pattern suggests that the main contribution of the Spatial Stage Summary is to organize spatial timeline evidence into a more explicit representation of object-state evolution, thereby supporting more reliable hidden-state reasoning in the final answer stage.

\subsubsection{Ablation of Spatial Summary Generation}

This ablation removes the spatial summary module, forcing the answering model to rely only on raw structured spatial evidence.

Table~\ref{tab:ablation_summary} shows that removing this module causes the largest performance drop among the tested components, consistently hurting all backbones across all benchmarks. The effect is most evident on LongVideoBench, where VideoLLaMA3-7B drops from 64.5 to 60.5 and Qwen2.5-VL-7B from 62.8 to 59.0. Similar degradations are also observed on MLVU and VideoMME. These results indicate that spatial summary generation is a critical interface between graph-structured memory and final answer prediction, enabling the model to convert low-level spatial and event records into an explicit account of object-state evolution.

% \subsubsection{Ablation of Key Clip Retrieval}

% This ablation removes final-stage key clip retrieval and instead feeds the entire video to the final answer model. It is designed to test whether retrieving event-relevant clips provides more effective support than directly exposing the answering model to the full video.

% Table~\ref{tab:ablation_retrieval} shows that removing key clip retrieval leads to a consistent performance drop across all tested backbones and benchmarks. The degradation is especially clear on LongVideoBench, where InternVL2.5-2B decreases from 56.3 to 53.2, Qwen2.5-VL-3B from 59.8 to 56.5, Qwen2.5-VL-7B from 62.8 to 59.4, and VideoLLaMA3-7B from 64.5 to 60.9. Similar drops are also observed on MLVU and VideoMME; for example, VideoLLaMA3-7B decreases from 77.2 to 75.0 on MLVU and from 74.1 to 72.1 on VideoMME with subtitles. This pattern suggests that the main contribution of key clip retrieval is to present the answering model with temporally focused, question-relevant evidence, rather than requiring it to identify the critical moments from the full video on its own.

\subsubsection{Ablation of Key Clip Retrieval}

This ablation removes final-stage key clip retrieval and instead uses the full video as input.

Table~\ref{tab:ablation_retrieval} shows that removing key clip retrieval consistently reduces performance across all backbones and benchmarks, although the degradation is smaller than that caused by removing spatial summary generation or disappearance-cause reasoning. The effect remains clear on LongVideoBench. VideoLLaMA3-7B drops from 64.5 to 62.7, while Qwen2.5-VL-7B drops from 62.8 to 61.2. MLVU and VideoMME also show consistent declines after key clip retrieval is removed. Overall, key clip retrieval helps concentrate evidence by directing the answering model toward the most relevant temporal context.

\subsection{Case Study}

Figure~\ref{fig:case_study} presents two hidden-state spatiotemporal reasoning examples using VideoLLaMA3-7B, involving occlusion and containment. In the first, the model must infer the color of a car temporarily hidden behind a black car; in the second, it must determine which cup contains the target nails after repeated exchanges.

The baseline fails in both cases, showing limited ability to preserve object states during invisibility. In contrast, StateTrace answers both correctly by maintaining state continuity and summarizing query-relevant evidence, demonstrating the importance of explicit state memory for reasoning under temporary invisibility.

% \section{Conclusion}

% In this work, we introduced \textit{StateTrace}, a novel object-centric framework for hidden-state spatiotemporal reasoning in long videos. Unlike conventional VideoLLMs that mainly rely on visible evidence at answer time, StateTrace builds a reusable spatiotemporal state memory that organizes object trajectories, inter-object relations, and visibility-transition events throughout the video. Based on this memory, StateTrace retrieves question-relevant state evidence and summarizes object-state evolution, enabling the answering model to reason more explicitly about why an object disappears, how its state evolves while invisible, and whether that state should still persist at query time. To better evaluate this capability, we further introduced HSR-Bench, a new diagnostic benchmark for hidden-state reasoning under occlusion, containment, covering, and long-delay state querying. Extensive experiments on both public long-video benchmarks and HSR-Bench demonstrate that StateTrace consistently improves strong VideoLLM backbones, with especially large gains on occlusion-centric hidden-state reasoning tasks. These results highlight the importance of explicit state-centric reasoning for long-video understanding and suggest a promising direction for future multimodal video reasoning systems.

\section{Conclusion}

We presented \textit{StateTrace}, an object-centric framework that builds reusable state memory for explicit reasoning over object disappearance, hidden-state evolution, and persistence. We also introduced HSR-Bench for evaluating occlusion-based hidden-state reasoning. Experiments on public benchmarks and HSR-Bench show consistent improvements over strong VideoLLMs, demonstrating the value of state-centric reasoning for long-video understanding.
\begin{acks}
This research is funded by the National Natural Science Foundation of China (No. 62536009 and No. 62406347).
\end{acks}

\bibliographystyle{ACM-Reference-Format}
\balance
\bibliography{references}
\end{document}